\documentclass[10pt,twocolumn,letterpaper]{article}

\usepackage[pagenumbers]{cvpr} % To force page numbers, e.g. for an arXiv version

\usepackage{graphicx}
\usepackage{amsmath}
\usepackage{amssymb}
\usepackage{booktabs}
\usepackage{multicol}
\usepackage{multirow}
\usepackage{array}
    \newcolumntype{P}[1]{>{\centering\arraybackslash}p{#1}}
    \newcolumntype{M}[1]{>{\centering\arraybackslash}m{#1}}

\usepackage[pagebackref,breaklinks,colorlinks]{hyperref}

\usepackage[capitalize]{cleveref}
\crefname{section}{Sec.}{Secs.}
\Crefname{section}{Section}{Sections}
\Crefname{table}{Table}{Tables}
\crefname{table}{Tab.}{Tabs.}

\def\cvprPaperID{*****} % *** Enter the CVPR Paper ID here
\def\confName{CVPR}
\def\confYear{2022}

\title{simCrossTrans: A Simple \textbf{Cross}-Modality \textbf{Trans}fer Learning for Object Detection with \textbf{C}onvNets or Vision \textbf{T}ransformers}

\author{Xiaoke Shen \\
Hunter College, CUNY \\
New York City, USA \\
{\tt\small xs54@hunter.cuny.edu}
\and
Ioannis Stamos \\
Hunter College \& The Graduate Center, CUNY \\
New York City, USA \\
{\tt\small istamos@hunter.cuny.edu}
}

\begin{document}
\maketitle
%%%%%%%%% ABSTRACT
\begin{abstract}
   Transfer learning is widely used in computer vision (CV), natural language processing (NLP) and achieves great success. Most transfer learning systems are based on the same modality (e.g. RGB image in CV and text in NLP). However, the cross-modality transfer learning (CMTL) systems are scarce. In this work, we study CMTL from 2D to 3D sensor to explore the upper bound performance of 3D sensor only systems, which play critical roles in robotic navigation and perform well in low light scenarios. While most CMTL pipelines from 2D to 3D vision are complicated and based on Convolutional Neural Networks (ConvNets), ours is easy to implement, expand and based on both ConvNets and Vision transformers(ViTs): 1) By converting point clouds to pseudo-images, we can use an almost identical network from pre-trained models based on 2D images. This makes our system easy to implement and expand. 2) Recently ViTs have been showing good performance and robustness to occlusions, one of the key reasons for poor performance of 3D vision systems. We explored both ViT and ConvNet with similar model sizes to investigate the performance difference. We name our approach \textbf{simCrossTrans}: \textbf{sim}ple \textbf{cross}-modality \textbf{trans}fer learning with \textbf{C}onvNets or Vi\textbf{T}s.\\ 
   Experiments on SUN RGB-D dataset show: with simCrossTrans we achieve $13.2\%$ and $16.1\%$ absolute performance gain based on ConvNets and ViTs separately. We also observed the ViTs based performs $9.7\%$ better than the ConvNets one, showing the power of simCrossTrans with ViT. simCrossTrans with ViTs surpasses the previous state-of-the-art (SOTA) by a large margin of $+15.4\%$ mAP50. Compared with the previous 2D detection SOTA based RGB images, our depth image only system only has a $1\%$ gap. The code, training/inference logs and models are publicly available at \url{https://github.com/liketheflower/simCrossTrans}.
\end{abstract}
%%%%%%%%% BODY TEXT
\section{Introduction}
\begin{figure}[t]
\begin{center}
   \includegraphics[width=1.0\linewidth]{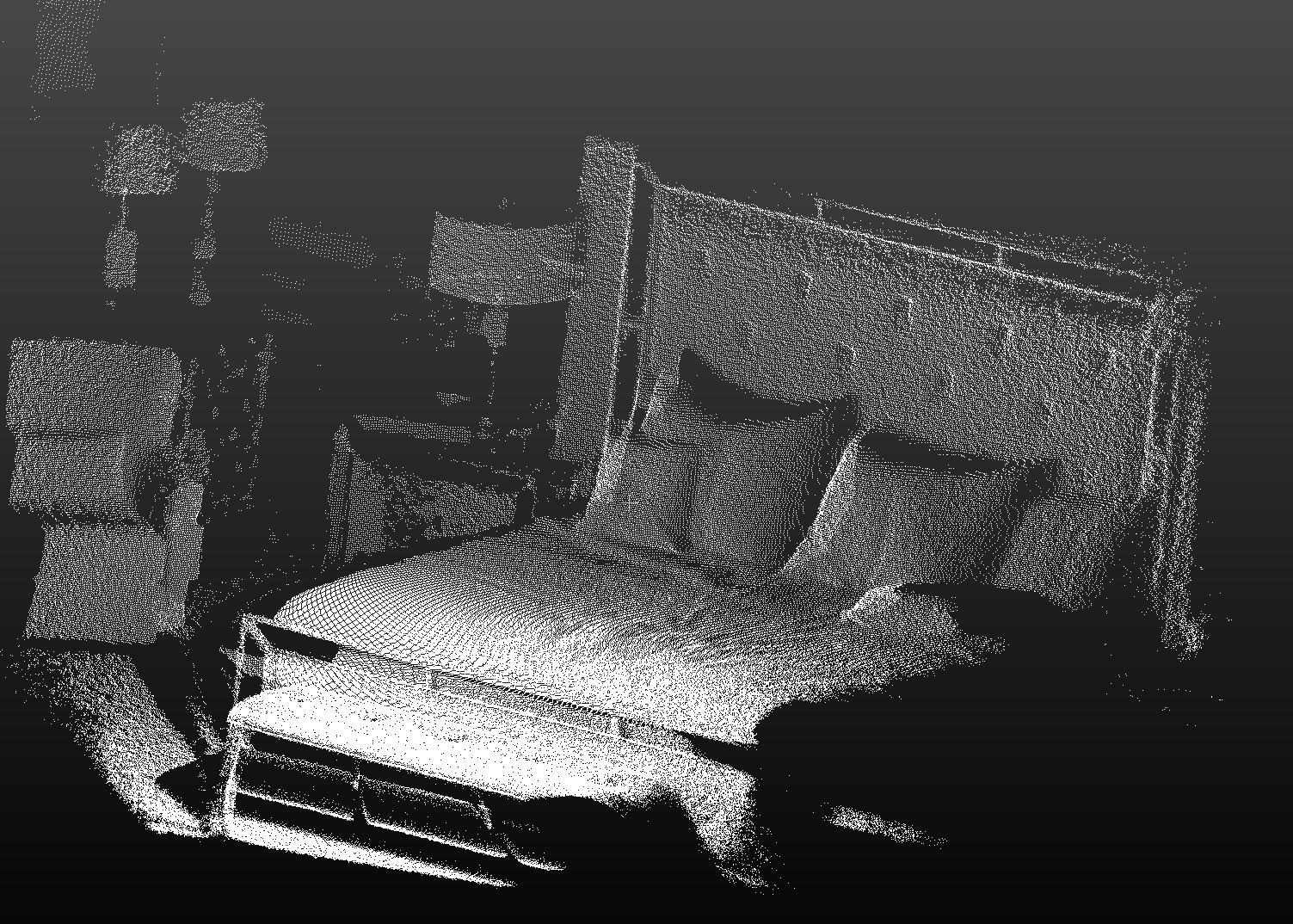}
\end{center}
    \caption{A cloud point from the SUN RGB-D dataset \cite{Song_2015_CVPR}.  }
\label{fig:cloud}
\end{figure}

\label{sec:intro}
\begin{figure*}[t]
\begin{center}
   \includegraphics[width=1.0\linewidth]{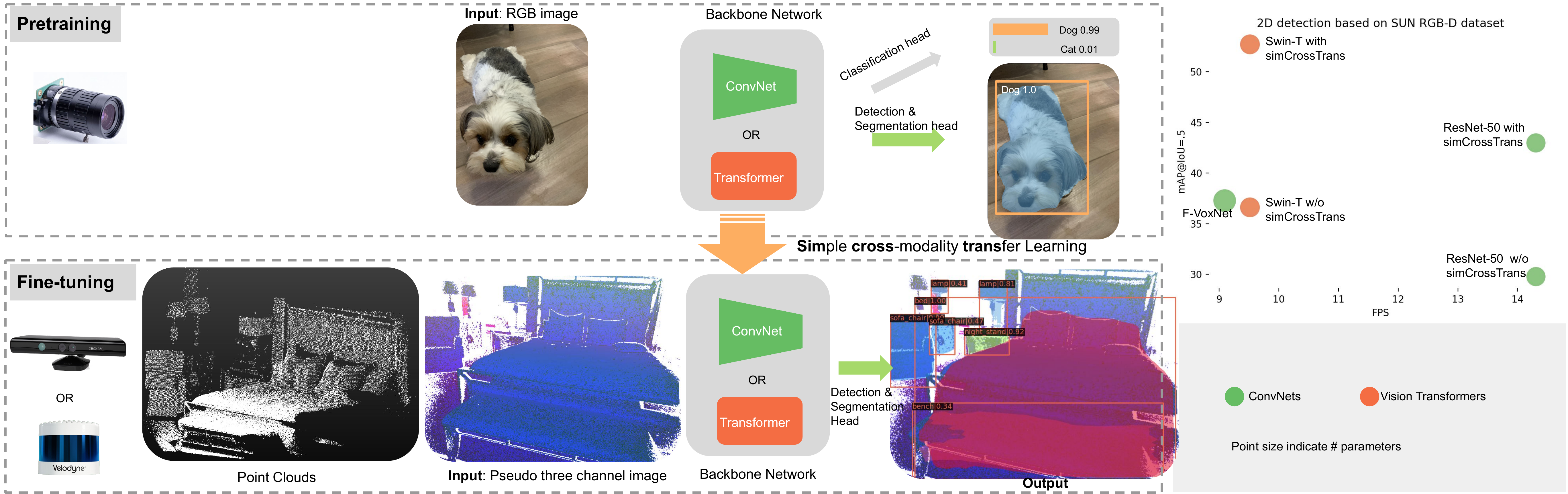}
\end{center}
    \caption{\textbf{Cross} Sensor/Modal \textbf{Trans}fer Learning approach: The top left one is a regular image classification, detection and segmentation pipeline with backbone network either ConvNets or Vision Transformers and head for each subtask. The pre-train is based on RGB color images collected by cameras. On the bottom left, the pre-trained model, including the backbone and heads, are fine tuned based the pseudo-images converted from point clouds, which can be collected by depth or Lidar sensors. The right shows clear performance boosting of with simCrossTrans V.S. without simCrossTrans based on the ConvNets (ResNet-50\cite{He_2016_CVPR}) or Transformers (Swin-T \cite{liu2021Swin}) as backbone network. Both ConvNets and Transformer surpass the previous state-of-the-art depth image only 2D detection system, Frustum Voxnet \cite{Shen_2020_WACV}, by a large margin. Details of the result see table \ref{2d_res}.}
\label{fig:pipeline}
\end{figure*}

Human beings (also some animals) have excellent vision systems by observing the colorful word using two eyes. Meanwhile, a stereoscopic vision system from two eyes makes human have an implicit depth sensor to infer the depth information to objects. We can build robots to have similar vision systems by applying 2D camera sensor and 3D Lidar or depth sensors. However, not all the scenarios, 2D cameras are available, such as night time or in some cases, people's privacy are being protected and hence that 2D cameras are not allowed to be deployed on robotics. This makes the 3D sensor only vision systems desirable for robotics. What's more, the 3D sensor only based on system can work well in the low light or no light scenarios, hence, it is environment friendly and cleaner. Thinking about how much electric bills can be saved if we have a warehouse robotic based on 3D sensors only and no need using lights. Motivated by this, we work on exploring the upper bound performance of the 3D sensor only vision systems.\\

When thinking about the human being's vision system, we observing the colorful 3D world since we first open our eyes. Unsurprisingly, when the first time a baby see a photo or picture, which is essentially a projection of the 3D world to 2D, it is not hard for the baby to recognize the interesting objects on the photo. Also, in Figure \ref{fig:cloud} before, it is a point cloud image from the SUN RGB-D dataset. Although this may be the first time that a person sees this kind image, it is not hard to recognize the bed, pillows and lamps in this image. Meanwhile, based on experience studying from real life, we can also infer some objects should have low possibility to be observed such as cars, trains, etc. Why it works? One possible reason is although the the madalities used to percept the world is different, if we observe the same world, the context or prior knowledge are sharing, hence can be transferred. The difference between human and robotics are: human is transferring knowledge learning from 3D world to 2D. For robotics, it will be easier if we do the reverse way due to there are significantly more 2D data and models trained based on those data available.\\
\begin{figure*}[t]
\begin{center}
   \includegraphics[width=1.0\linewidth]{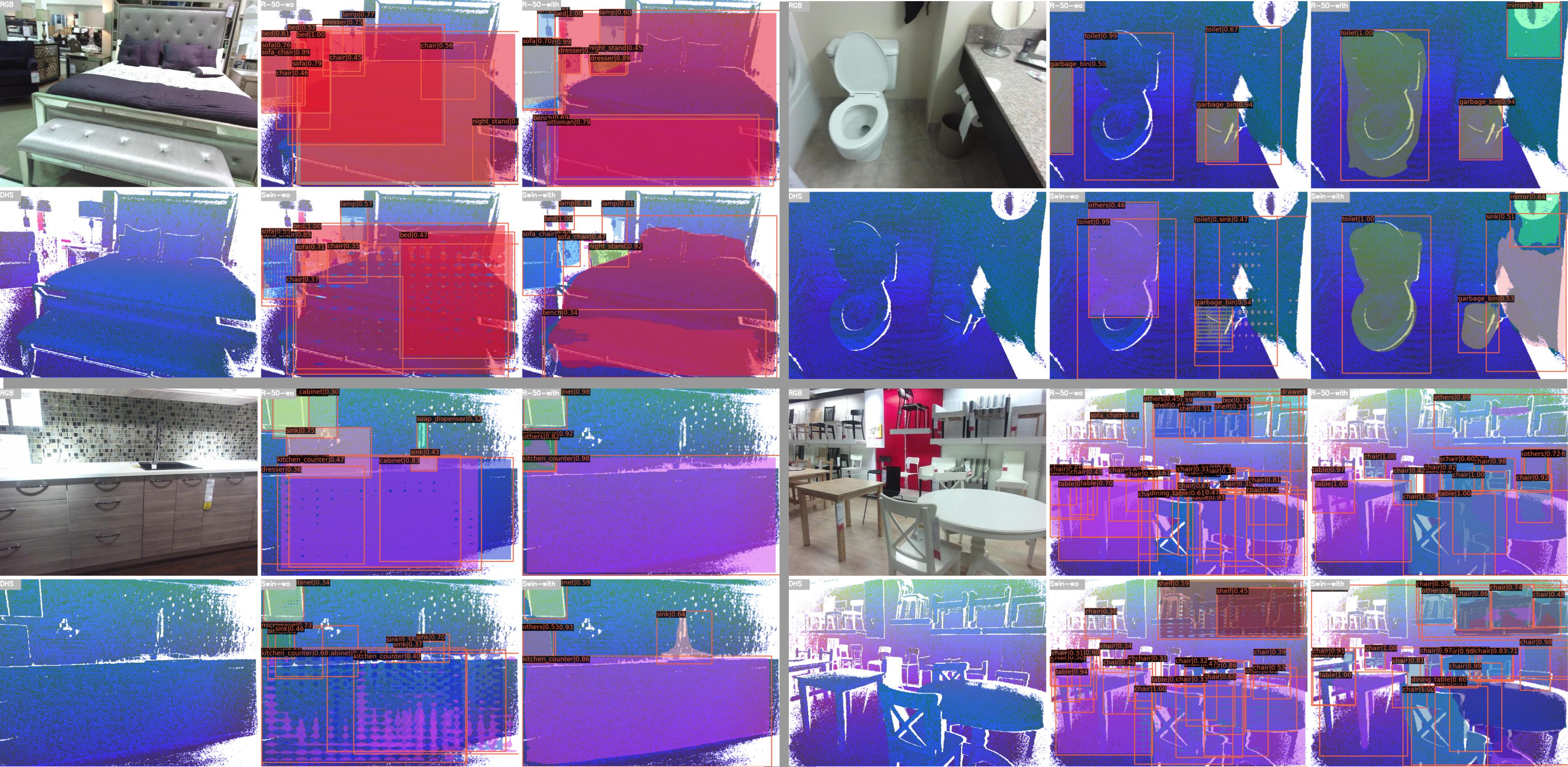}
\end{center}
    \caption{2D detection results based on the SUN RGB-D validation data split. It has 2D detection based on 4 images as shown in the top left, top right, bottom left and bottom right blocks. Each block has 9 images: top left is the original RGB image, it is only used for better observe the objects, the RGB images are not used during the training and testing process. bottom left is the pseudo 2D image converted from point cloud. The top middle is the ResNet-50 backbone based 2D detection result without using simCrossTrans and the top right is with using simCrossTrans. The bottom middle is the Swin-T backbone based 2D detection result without using simCrossTrans and the bottom right is with using simCrossTrans. More results can be found in \url{https://youtu.be/qQ0w-GpPzjI}}
\label{fig:more_vis}
\end{figure*}
How can we transfer the knowledge from 2D sensor to 3D sensor? The answer is quite simple: using transfer learning. Transfer learning has been widely used in CV and NLP and achieve great success. Originally, the transfer learning in CV is following a supervised pretraining approaches, such as pretraining on ImageNet and fine-tune on other datasets for image classifications or object detection and instance segmentation \cite{He_2016_CVPR}. At the same time, NLP also achieves a tremendous success by applying self-supervised pretraining approaches to further improve the following subtasks' performance. Classical works are \cite{devlin-etal-2019-bert}. Inspired by the success of self-supervised pretraining approaches, CV researchers are also trying to use self-supervised pretraining approaches \cite{DBLP:journals/corr/abs-2111-06377} to build better transfer learning systems. However, all those systems mentioned here are based on the same modality. Whether we can achieve similar success when applying cross-madality transfer learning, which is pretraining a model based on one madality and fine-tune the model based on another modality? The cross-modality transfer learning (CMTL) work are rare compared with the transfer learning between the same modality. There are more general CMTL system, which pretrains a model from one modality and fine tune a model from a totally different model, such as from text to vision shown in work \cite{10.1145/3464324} and from vision to sound shown in \cite{10.1007/978-3-030-91581-0_32}. This is not the scope of this work due to: our goal is investigating the capability of cross modality transfer learning from 2D to 3D sensors and we need the different modality sharing the same observing target to make sure the knowledge relatively easier to be transferred.\\
Inspired by the ViT \cite{dosovitskiy2021an} work, where the authors built a image classification system with as less modification to the original transformer \cite{transformer_cite} framework as possible to build a unified model which can be used to process vision and languages, we propose a framework with almost zero modification to any 2D object detection based system by adjusting the raw point cloud data to 3-channel pseudo-images (See the bottom left two images as example, the first one is the same point cloud image as Figure \ref{fig:cloud} and the second is the converted 3 channel pseudo image). By doing this, we have at least two benefits: 1) simple: almost no need to modify the model architecture, we can train the model based on new modality. 2) easy to be expanded: any improvement in regular 2D vision system can be expanded to further improve our 3D system by following our approach.\\

Due to those two advantages mentioned above, we can easily explore the latest progress in vision which is Vision Transformer based models, specifically, it is the Swin-Transformer \cite{liu2021Swin}. Self-attention-based architectures, in particular Transformers \cite{transformer_cite}, have become the model of choice in NLP and achieved great success with transfer learning. The ViT work cuts the image into patches and project each patch's raw pixels to a vector, and the vector can be treated as the word embedding in NLP. Hence the image can be fed into a standard transformer to performance image classification task. With enough pretraining data, the ViT suppass the SOTA on image classification based on ImageNet \cite{deng2009imagenet} dataset. The swin-transformer \cite{liu2021Swin} further extend the ViT to solve the object detection and instance segmentation tasks in CV by introducing the hierarchical transformer to extract different scale features and shifted window approach approaches instead of sliding window to reduce the computation complexity. The Swin-Transfomer achieved great success and provides the CV an alternative approach besides of using ConvNets. Although the Swin-Transformer performances well based 2D RGB images, why it works better than ConvNets are not very clear. Besides that, whether it can also achieve a better performance with CMTL is not verified. However, from \cite{naseer2021intriguing} it shows that ViT includes impressive robustness to severe occlusions, which is one of the critical causes for the poor performance of many 3D vision systems. This observation encourages us to explore the ViT's performance. Based on above, in this work, we explored both ConvNets\cite{10.1162/neco.1989.1.4.541} and ViTs, and provides detailed comparison and analysis.\\

The workflow for our system is pretty straight forward: pretrain based on RGB images and then fine-tune based on the 3 channel pseudo images converted from point cloud. An illusttration can be found in Figure \ref{fig:pipeline}. We name our system as \textbf{simCrossTrans}: \textbf{sim}ple \textbf{cross}-modality \textbf{trans}fer learning with \textbf{C}onvNets or Vi\textbf{T}s with respect to both ConvNets and Transformers.\\

%For the CMTL based on different vision modalities, 
%Based on this, we propose a cross-modality transfer learning system.\\
%\\and well pretrained models based on 2D images.

%are the same, although the  
%Suppose that a person lost his/her vision one day, instead he/she acquires a computer 3D vision system. It will not be hard for him/her to infer there is a car in front of him/her based on his/her previous experience. Also, it will not be hard for him/her to infer that the person shape object in the driven seat should be a person instead of a snow man based on his prior knowledge. Inspired by this, we argue that although sensors are different, if they are used to observe the same world, the knowledge learned from one sensor can be transferred to another sensor. Based on this we propose an \textbf{Cross} Sensor \textbf{Trans}fer learning system: \textbf{simCrossTrans}. \\
%Exploring the performance of Cross sensor transfer learning is meaningful due to following reasons. First, (Data matters) the recent significant performance. Second, not all the sensors can collect similar quality data due to the availabilities of this sensor and other reasons such as cost. \\
% Transfer learning achieve great success and we want to see how it works.

%Based on those motivitations, w
In summary, in this article, we want to find answers for the following questions:\\
\begin{itemize}
    \item Whether we can achieve performance boosting for simCrossTrans based on vision transformers (We did not mention ConvNets, as the performance boosting was observed in Frustum-VoxNet \cite{Shen_2020_WACV}).
    \item If simCrossTrans works, why?
    \item For ConvNets and Vision Transformers, which can achieve a better performance when applying simCrossTrans? Why one is better than the other?
\end{itemize}
Through experiments, we indeed have interesting observations and achieve SOTA results on 2D object detection based on point cloud only images. In Figure \ref{fig:more_vis} we show some visualization of the 2D object detection by comparing of with and without the simCrossTrans based on ResNet-50 and Swin-T as backbone network. From the result, we can see by using the simCrossTrans, not only we achieve significantly bounding box detection performace boosting but also the system with simCrossTrans can produce reasonable segmentation outputs even without fine tuning on that branch. We will explore details in the following session. Here are our contributions:\\
\begin{itemize}
    \item We highlight the performance boosting of using simCrossTrans based on ConvNets.
    \item We experimentally show the performance boosting of using simCrossTrans based on vision Transformers.
    \item We comprehensively compare the performance difference of applying simCrossTrans between ConvNets and Vision Transformers to provide insights about the difference of those two backbone networks.
    \item When applying simCrossTrans on a vision transformer based detection framework, we achieve SOTA on 2D object detection based on point clouds only.
    %\item By further feeding our improved 2D detection results to 2D driven 3D detection systems, we can harvest a better 3D detection outputs. This is critical and can benefit 3D sensor only object detection systems. (TOBE ADDED)
    \item We open source our code, training/testing logs and model checkpoints to benefit the vision community.
\end{itemize}
% we tried. based on convnet and transformers. Both works.
% Our contributions can be list as following:
%The solutions, based on autoregressive language modeling in GPT [47, 48, 4] and masked autoencoding in BERT [14], 
% why it is interesting
%

%-------------------------------------------------------------------------
\section{Related Work}
\textbf{Transfer learning with same modality.} Transfer learning is widely used in computer vision (CV), natural language processing (NLP) and biochemistry. Most transfer learning systems are based on the same modality (e.g. RGB image in CV and text in NLP). For the CV, common transfer learning is based on supervised way such as works in R-CNN \cite{DBLP:journals/corr/GirshickDDM13}, Fast RCNN \cite{DBLP:conf/iccv/Girshick15}, Faster RCNN \cite{DBLP:conf/nips/RenHGS15}, FPN \cite{Lin_2017_CVPR}, mask R-CNN \cite{maskrcnn}, YOLO\cite{DBLP:journals/corr/RedmonDGF15}, YOLO9000\cite{DBLP:journals/corr/RedmonF16}, RetinaNet \cite{DBLP:journals/corr/abs-1708-02002} use a pretrained backbone network model based on ImageNet classification task and the model is further trained based on the following task datasets such as COCO to achieve object detection or/and instance segmentation tasks. In the NLP, the transfer learning such as BERT \cite{devlin-etal-2019-bert}, GPT\cite{gpt1}, GPT-2\cite{gpt2}, GPT-3\cite{gpt3} are mainly based on self-supervised way and achieve great success. Inspired by the success of the self-supervised way tansfer learning, the CV community is also exploring the self-supervised way to explore new possibilities, one recent work which is similar to the BERT in NLP is MAE \cite{DBLP:journals/corr/abs-2111-06377}. The MolGNN \cite{molgnn1, liu2021covid} in bioinformatics use a self-supervised way based on Graph Neural network (GNN) in the pretraining stage and achieve good performance in a few shot learning framework for the following subtasks. For this work, we explore the cross modalitiy transfer learning from a pretrained model under the supervised learning approach.\\

\textbf{Cross sensor/modality transfer learning.} In this work, we focus on the vision related cross modality transfer learning. The cross modality between vision and other modalities, such as from text to vision shown in work \cite{10.1145/3464324} and from vision to sound shown in \cite{10.1007/978-3-030-91581-0_32} is beyond the scope of this work and will not be discussed. \cite{Kim2017ModalitybridgeTL} explored the cross modality transfer learning from RGB to X-ray images, however, this work was mainly focusing on the image classification and it is based on ConvNets only. \cite{cross_organ} use  pre-train ConvNets (VGG16 and ResNet 16) on an unrelated set of medical images (mammograms) first and then fine-tune on brain magnetic resonance (MR) images to address segmentation task. This work is mainly focusing on segmentation and transformer is not explored. \cite{s21237950} change the one channel depth image into 3 channels and use two separate VGG16 based ConvNets for converted 3 channel depth images and RGB images to extract features. The extracted features are combined later to solve the scene classification task. This framework is complicated due to two backbone networks and post combining networks have to be used. Also, this system does not support the uni-modality inference scenario.  Frustum-Voxnet \cite{Shen_2020_WACV} used pretrained weights from the RGB images to fine tune the point cloud converted pseudo image, which is most close to this work. However, in \cite{Shen_2020_WACV}, the performance difference of using CMTL and without using the CMTL are not compared. Plus, the work in \cite{Shen_2020_WACV} is based on ConvNets, we further explored the performance difference of using ViTs and also have a deep analysis about the reasons.\\

\textbf {Projecting 3D sensor data to 2D Pseudo Images.} There are different ways to project 3D data to 2D features. HHA was proposed in \cite{GuptaGAM14} where the depth image is encoded with three channels: Horizontal disparity, Height above ground, and the Angle of each pixel’s local surface normal with gravity direction. The signed angle feature described in\cite{6375012} measures the elevation of the vector formed by two consecutive points and indicates the convexity or concavity of three consecutive points. Input features converted from depth images of normalized depth(D), normalized relative height(H), angle with up-axis(A), signed angle(S), and missing mask(M) were used in \cite{7785116}. DHS images are used in \cite{Shen_2020_WACV, Shen_2020_v2}. 

%\textbf{Object Detection Based on Pseudo Images from 3D Sensor}\\

\textbf{Object Detection Based on RGB images by ConvNets}\\
RGB-based approaches can be summarized as two-stage frameworks (proposal and detection stages) and one-stage frameworks (proposal and detection in parallel). Generally speaking, two-stage methods such as R-CNN \cite{DBLP:journals/corr/GirshickDDM13}, Fast RCNN \cite{DBLP:conf/iccv/Girshick15}, Faster RCNN \cite{DBLP:conf/nips/RenHGS15}, FPN \cite{Lin_2017_CVPR} and mask R-CNN \cite{maskrcnn} can achieve a better detection performance while one-stage systems such as YOLO\cite{DBLP:journals/corr/RedmonDGF15}, YOLO9000\cite{DBLP:journals/corr/RedmonF16}, RetinaNet \cite{DBLP:journals/corr/abs-1708-02002} and FFESSD \cite{app9204276} are faster at the cost of reduced accuracy. For deep learning based systems, as the size of network is increased, larger datasets are required. Labeled datasets such as PASCAL VOC dataset \cite{pascal-voc-2012} and COCO (Common Objects in Context) \cite{DBLP:journals/corr/LinMBHPRDZ14} have played important roles in the continuous improvement of 2D detection systems. Most systems introduced here are based on ConvNets. Nice reviews of 2D detection systems can be found in \cite{DBLP:journals/corr/abs-1905-12683}. \\
\textbf{Object Detection Based on RGB images by Vision Transformers}\\
When replacing the backbone network from ConvNets to Vision Transformers, the systems will be adopted to Vision Transformers backbone based object detection systems. The most successfully systems are Swin-transformer\cite{liu2021Swin} and Swin-transformer v2 \cite{DBLP:journals/corr/abs-2111-09883}.\\

\section{Approach}
\subsection {Convert point clouds to pseudo 2D image}
In order to use pretrained models based on RGB images, we convert point clouds to pseudo 2D images with 3 channels. The point clouds can be converted to HHA or any three channels from DHASM introduced in \cite{allancnn}.\\

For this work, we follow the same approaches in Frustum VoxNet\cite{Shen_2020_WACV} by using DHS to project 3D depth image to 2D due to: 1) \cite{allancnn} shows DHS can provide a solid result; 2) have a fair comparison with the Frustum VoxNet\cite{Shen_2020_WACV} work.\\

Here is a summary of the DHS encoding method: Similar to\cite{GuptaGAM14,allancnn}, we adopt \textbf{D}epth from the sensor and \textbf{H}eight along the sensor-up (vertical) direction as two reliable measures. \textbf{S}igned angle was introduced in \cite{OnlineClassificationStamos}: denote $X_{i,k} = [x_{ik},y_{ik},z_{ik}]$ the vector of 3D coordinates of the $k$-th point in the $i$-th scanline. Knowledge of the vertical direction (axis $\mathbf{z}$) is provided by many laser scanners, or even can be computed from the data in indoor or outdoor scenarios (based on line/plane detection or segmentation results from machine learning models) and is thus assumed known. Define $D_{i,k} = X_{i,k+1} - X_{i,k}$ (difference of two successive measurements in a given scanline $i$), and $A_{ik}$: the angle of the vector ${D_{i,k}}$ with the pre-determined $\mathbf{z}$ axis (0 to 180 degrees). The \textbf{S}igned angle $S_{ik} =sgn(\mathbf{D_{i,k}}\cdot \mathbf{D_{i,k-1}})*A_{ik}$: the sign of the dot product between the vectors $D_{i,k}$ and $D_{i,k-1}$, multiplied by $V_{ik}$. This sign is positive when the two vectors have the same orientation and negative otherwise. Those 3 channel pseudo images are normalized to 0 to 1 for each channel. Some samples DHS images can be seen in Figure \ref{fig:pipeline} and \ref{fig:more_vis}.\\
\\
\subsection{The Simple Cross-Modality transfer learning}
% TOBE ADDED
The \textbf{sim}ple \textbf{cross}-modality \textbf{trans}fer learning approach contains 3 key steps: 1) Pre-train 2D vision system based on 2D RGB images in either supervised way or self-supervised way. In this work, we explore the supervised approach; 2) Convert the point clouds to pseudo images. 3) Fine-tune the pretrain model from the RGB image based on the pseudo images. An illustration of this pipeline is shown in Figure \ref{fig:pipeline}. Both \textbf{C}onvetNets and Vision \textbf{T}ransformers can be used as backbone networks.\\
\section{SUN RGB-D dataset Experiments}

\begin{table*}
\begin{center}
	\resizebox{\textwidth}{!}{
\begin{tabular}{|c|l|l|c|c|c|c|c|c|c|c|c|c|c|c|c|c|c|c|c|c|}
\hline
Image Source &Methods	&Backbone& bed     &toilet&  \shortstack{night \\stand}&     bathtub&        chair  &dresser &sofa& table & desk&	bookshelf	&\shortstack{sofa\\ chair}&\shortstack{kitchen\\ counter}&\shortstack{kitchen\\ cabinet}&\shortstack{garbage\\ bin}&microwave&sink&\shortstack{SUNRGBD10\\$mAP_{50}$}&\shortstack{SUNRGBD16\\$mAP_{50}$}\\
\hline
RGB $\&$Depth&RGB-D RCNN\cite{GuptaGAM14}&VGG&76.0&   69.8&37.1&   49.6&   41.2   &31.3&42.2&43.0& 16.6&   34.9 &N/A&N/A&N/A&46.8&N/A&41.9&44.2&N/A\\
\hline\hline
	%\shortstack{Frustum PointNets\cite{Qi_2018_CVPR}\\(RGB)}&56.7&   43.5&57.4&49.9&	77.8&	67.2&	33.3&	37.2&	81.3&	64.1&N/A&N/A&N/A&N/A&N/A&N/A\\
  \multirow{3}{*}{RGB}&      2D-driven\cite{Lahoud_2017_ICCV}&VGG-16&74.5&  86.2& 49.5&   45.5&   53.0&29.4&49.0&42.3&	22.3&	45.7&	N/A&N/A&N/A&N/A&N/A&N/A&49.7&N/A\\
&  	Frustum PointNets\cite{Qi_2018_CVPR}&adjusted VGG from SSD&56.7&   43.5& 37.2&  \textbf{ 81.3}&   64.1&\textbf{33.3}&57.4&\textbf{49.9}&	\textbf{77.8}&	\textbf{67.2}&	N/A&N/A&N/A&N/A&N/A&N/A&\textbf{56.8}&N/A\\
&	F-VoxNet\cite{Shen_2020_WACV}&ResNet 101	&81.0&  \textbf{89.5}& 35.1&   50.0&   52.4&21.9&53.1&37.7	&18.3	&40.4	&47.8&22.0&29.8&52.8&39.7&31.0&47.9&N/A\\
\hline\hline
\multirow{5}{*}{Point Cloud only}& 	F-VoxNet\cite{Shen_2020_WACV}&ResNet 101&78.7 & 77.6& 34.2&   51.9  &51.8&16.5&48.5	&34.9	&14.2	&19.2 &48.7&19.1&18.5&30.3&22.2&30.1&42.8&37.3\\
%	\hline\hline
	&Ours w/o simCrossTrans&ResNet 50&	72.2&	51.7	&22.0&	44.4&	49.5	&9.6	&33.6	&33.5&	9.7&	12.2	&41.8&	15.7&	25.3&22.8	&12.9&	19.5&33.8	&29.8\\
&		Ours with simCrossTrans&ResNet 50&82.1	&78.8&	43.5	&49.9&	60.0	&14.8	&50.3	&41.1	&13.0	&31.6	&58.9	&26.7	&41.4	&25.6&	30.6	&39.3&46.5&\textbf{43.0} \\
	&	Ours w/o simCrossTrans&Swin-T&76.7	&58.4	&23.1	&49.7	&57.2	&12.1	&40.5	&41.5	&13.5&	22.8&	54.7&	19.7&	37.8&	29.2&	23.7&	25.2&39.6&36.6\\
&	Ours with simCrossTrans& Swin-T&\textbf{87.2}&	87.7&	\textbf{51.6}	&69.5	&\textbf{69.0}	&27.0	&\textbf{60.5}	&48.1&	19.3&	38.3&	\textbf{68.1}&	\textbf{30.7}&	\textbf{61.2}&	\textbf{35.5}&	\textbf{41.9}&	\textbf{47.7}&55.8&\textbf{52.7}\\
	%OURS(RGB-D)&59.7&  55.0&7.0&    46.5&   37.7&4.7&22.7&25.8	&4.7	&11.1	&32.0&16.2&16.9&28.7&10.2&17.7\\
\hline
\end{tabular}}
\end{center}
	\caption{2D detection results based on SUN-RGBD validation set. All systems are evaluated based on Depth Images. Evaluation metric is average precision with 2D IoU threshold of 0.5.}
	\label{2d_res}
\end{table*}

\subsection{SUN RGB-D dataset}
SUN RGB-D\cite{Song_2015_CVPR} dataset is an indoor dataset which provides both the point cloud and RGB images. In this work, since we are building a 3D only object detection system, we only use the point clouds for fine tuning. The RGB images are not used during the fine tuning process. For the point clouds, they are collected based on 4 types of sensors: Intel RealSense, Asus Xtion, Kinect v1 and Kinect v2. The first three sensors are using IR light pattern. The Kinect v2 is based on time-of-flight. The longest distance captured by the sensors are around 3.5 to 4.5 meters.\\

SUN RGB-D dataset splits the data into a training set which contains 5285 images and a testing set which contains 5050 images. For the training set, it further splits into a training only, which contains 2666 images and a validation set, which contains 2619 images. Similar to \cite{DBLP:journals/corr/SongX15,Lahoud_2017_ICCV,Shen_2020_WACV,Shen_2020_v2} , we are fine-tuning our model based on the training only set and evaluate our system based on the validation set. We call the only training dataset as train2666 in the future description.\\

\subsection{2D detection framework}
For the 2D detection, we use the classical object detection framework: Mask R-CNN  \cite{maskrcnn} in mmdetection \cite{mmdetection}. \\
\subsection{2D detection backbone networks}

\begin{table*}[h]
\scriptsize
\begin{center}
               \resizebox{1.0\linewidth}{!}{
\begin{tabular}{|c|c|ccc|ccc|ccc|ccc|ccc|}
\hline
     \multirow{2}{*}{Method}  & \multirow{2}{*}{Backbone Network}& \multicolumn{3}{c|}{SUNRGBD10}&\multicolumn{3}{c|}{SUNRGBD16}&\multicolumn{3}{c|}{SUNRGBD66}&\multicolumn{6}{c|}{SUNRGBD79}\\
    & &AP&AP$_{50}$&AP$_{75}$&AP&AP$_{50}$&AP$_{75}$&AP&AP$_{50}$&AP$_{75}$&AP&AP$_{50}$&AP$_{75}$&AP$_S$&AP$_M$&AP$_L$\\
\hline
      \hline
      w/o simCrossTrans & ResNet-50 & 15.6 &33.8  &11.4 &13.5&29.8&9.8&5.0&11.4&3.4&4.2&9.6&2.8&0.2&0.9&5.3\\
      with simCrossTrans & ResNet-50& 26.4 &46.5  &27.7 &24.0&43.0&24.9&10.3&19.4&10.0&8.8&16.5&8.5&0&3.2&10.7\\
      w/o  simCrossTrans & Swin-T& 18.4 &39.6  &14.6 &17.0&36.6&13.4&6.4&14.1&4.9&5.4&11.9&4.1&0.1&1.6&6.6\\
      with simCrossTrans & Swin-T& 33.3 &55.8  &34.7 &30.7&52.7&31.5&14.3&26.1&14.0&12.0&22.1&11.7&0.6&4.7&15.2\\
\hline
\end{tabular}}
\end{center}
\caption {More results comparison based on AP@IoU = .75, AP and AP of different scales.}
\label{more_2d_res}
\end{table*}

\begin{table}[h]
\scriptsize
\begin{center}
               \resizebox{1.0\linewidth}{!}{
%   \resizebox{1.0\textwidth}{!}{
                               %\resizebox{\textwidth}{!}{
\begin{tabular}{|c|c|c|c|c|c|}
\hline
      Method & Backbone Network& \# Parameters (M) & GFLOPs &Inference Time (ms) & FPS\\
\hline\hline
      F-VoxNet \cite{Shen_2020_WACV} & ResNet-101 & 64 &-& 110 & 9.1\\
      \hline
      simCrossTrans(ours) & ResNet-50 & 44 & 472.1&70 & 14.3\\
       simCrossTrans(ours) & Swin-T & 48 & 476.5&105 & 9.5\\
\hline
\end{tabular}}
\end{center}
\caption {Number of parameters and inference time comparison between Frustum VoxNet and ours based on ResNet 50 and Swin-T backbone networks with Mask R-CNN framework. All speed testing are based on a standard single NVIDIA Titan-X GPU.  }
\label{infer_time_comp_}
\end{table}

For the backbone network, we tried the 50 layer ConvNets: ResNet-50 \cite{He_2016_CVPR} and ViT based Swin-T \cite{liu2021Swin}. Similar to \cite{liu2021Swin}, we compare the ConvNets with ViT by using ResNet-50 and Swin-T due to they have similar number of parameters (see Table \ref{infer_time_comp_}) to have a relatively fair comparison.\\

\subsection{Pre-train}
Both the Swin-T and ResNet-50 based networks\footnote{The pretrained weights are loaded from  mmdetection \cite{mmdetection}.} are firstly pre-trained on ImageNet \cite{deng2009imagenet} and then pre-trained on the COCO dataset \cite{DBLP:journals/corr/LinMBHPRDZ14}.\\

\textbf{Data augmentation} When pre-training on COCO dataset, the images augmentation are applied during the training stage by: randomly horizontally flip the image with probability of 0.5; randomly resize the image with width of 1333 and height of several values from 480 to 800 (details see the configure file from the github repository); randomly crop the original image with size of 384 (height) by 600 (width) and resize the cropped image to width of 1333 and height of several values from 480 to 800.\\

\subsection{Fine-tuning}
\textbf{Data augmentation}: We follow the same augmentation with pre-train stage. The raw input images has the width of 730 and height of 530. Those raw images are randomly resized and cropped during the training. During testing, the images are resized to width of 1120 and height of 800 which can be divided by 32.\\

\textbf{Hardware:} For the fine-tuning, we use a standard single NVIDIA Titan-X GPU, which has 12 GB memory. We fine-tune the Resnet-50 and Swin-T backbone network for 100 epochs\footnote{For ResNet-50, as the dataset is repeated 3 times for each epoch, so in total it has 399 iterations for 100 epochs. For Swin-T, the dataset is not repeated, in total it has 133K iterations for 100 epochs.}. It took 16 hours to train the ResNet-50 based network with batch size of 2 (for 133 iterations) and 29 hours for Swin-T based network with batch size of 2 (for 133K iterations).\\

\textbf{Fine-tuning subtasks:} We focus on the 2D object detection performance, so we fine-tune the model based on the 2D detection related labels. We kept the mask branch without training to further verify whether reasonable mask detection can be created by naively apply a model learned from the RGB images.\\
\subsection{Results}
\begin{figure}[t]
\begin{center}
   \includegraphics[width=1.0\linewidth]{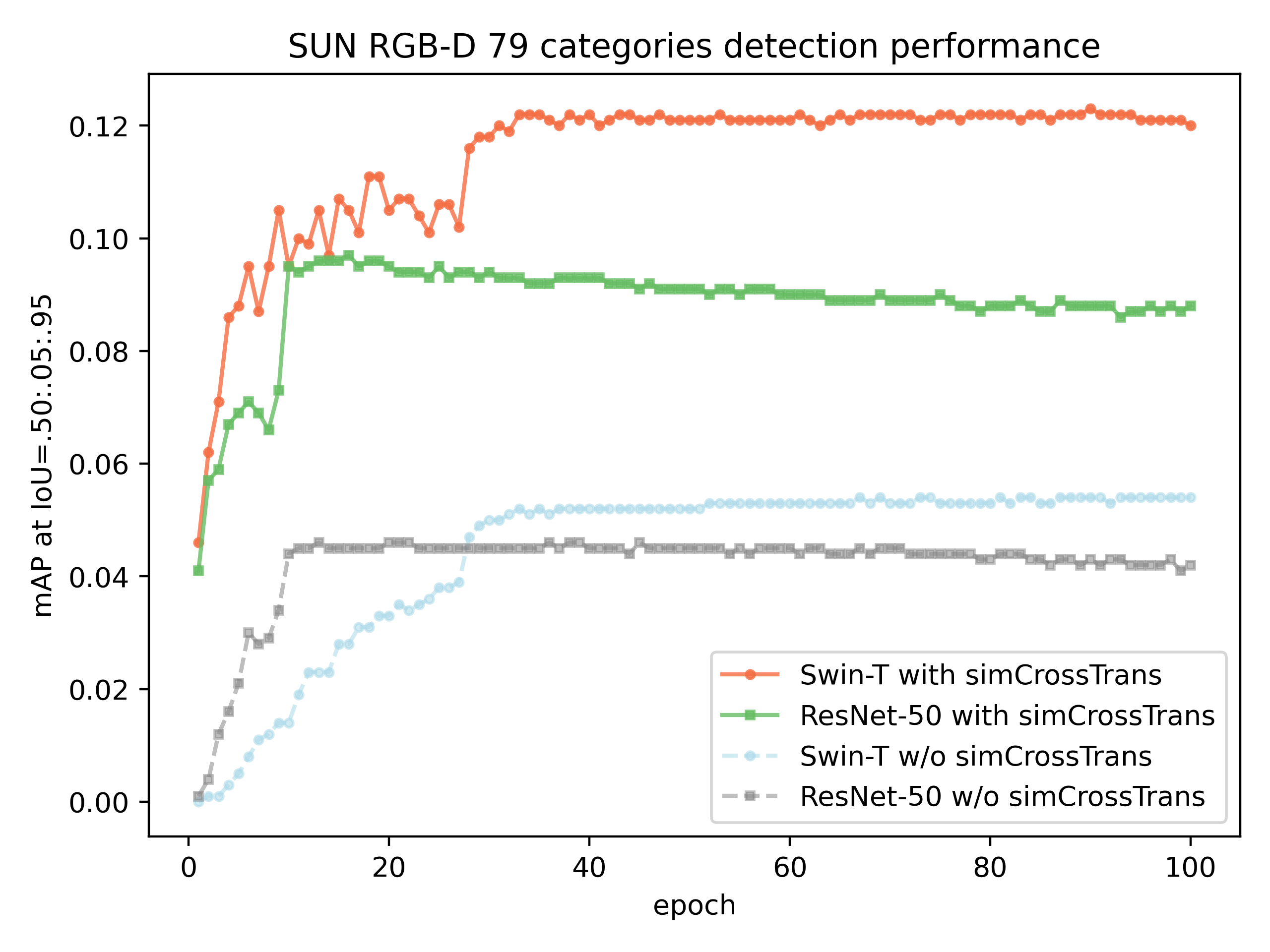}
\end{center}
    \caption{The updating the mAP based on SUNRGBD79 with fine tuning going on.}
\label{fig:mAP_sunrgbd79}
\end{figure}

\begin{figure*}[t]
\begin{center}
   \includegraphics[width=1.0\linewidth]{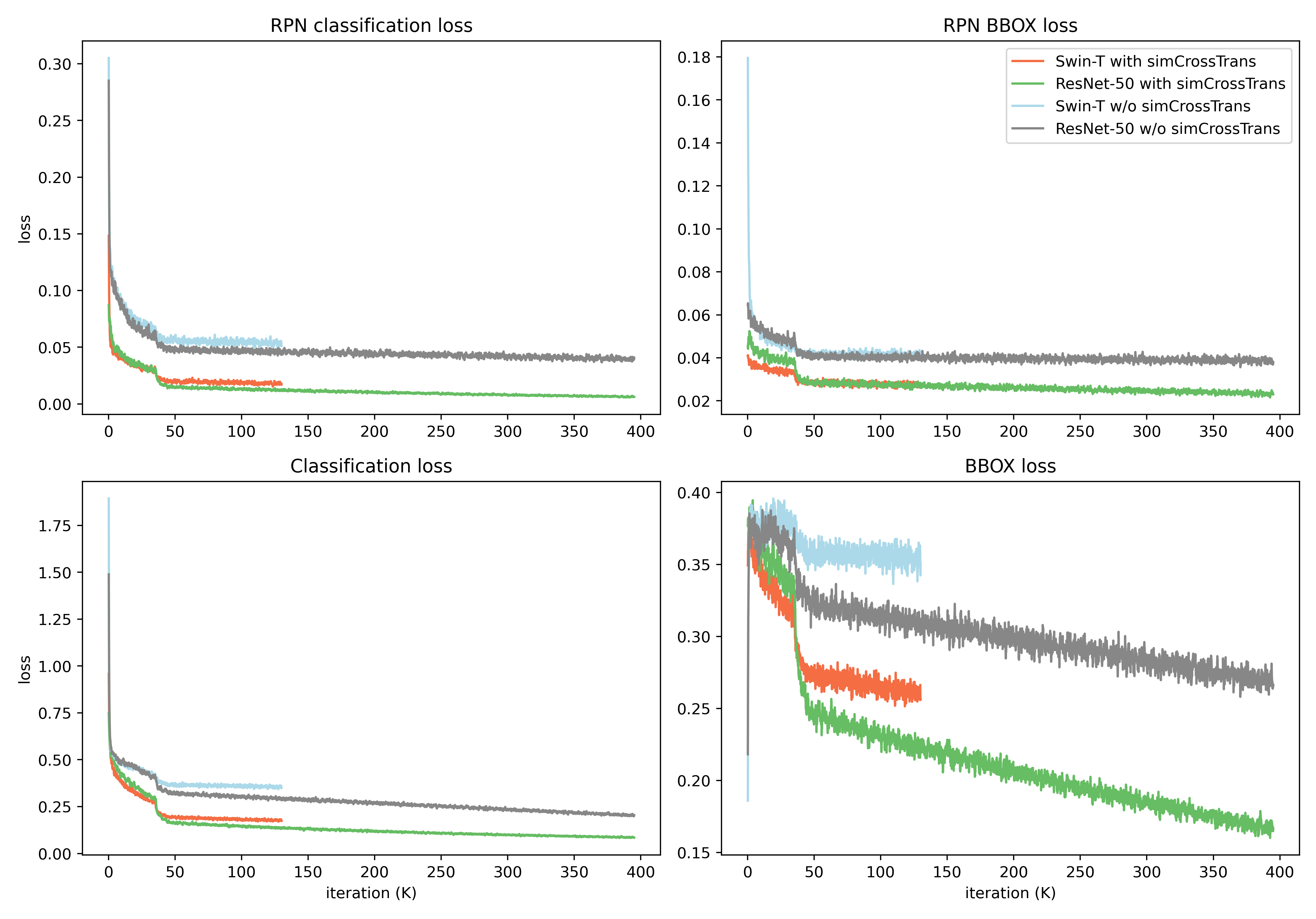}
\end{center}
    \caption{Train loss for based on different components. As the Mask-RCNN is a two stage detection network, it has the proposal stage and detection stage. For the proposal stage, the Region Proposal Network (RPN)'s classification loss and bounding box loss are shown in top. The detection stage, the classification and bounding box loss based on SUNRGBD79 categories are shown in bottom. }
\label{fig:losses}
\end{figure*}
%A comprehensive comparison based on the AP at the IoU=0.5 are shown in the table \ref{2d_res}. More detailed descriptions can be found in the following sub-sessions.\\
\textbf{Evaluation metrics:} Following the previous works mentioned in \ref{2d_res}, we firstly use the AP50: Average Precision at IoU = 0.5 as evaluation metric. We also use the COCO object detection metric which is AP75: Average Precision at IoU = 0.75 and a more strict one: AP at IoU = .50:.05:.95 to evaluate the 2D detection performance. \\

\textbf{Evaluation subgroups.} SUN RGB-D has about 800 2D objects. Not all the categories are detected in the previous works. In order to compare with different previous works, we group the subgroup to SUNRGBD10, SUNRGBD16 which have 10 categories and 16 categories. We also introduce SUNRGBD66 and SUNRGBD79 which contains 66 and 79 categories. The 66 categories are relatively easier to be recognized from the raw cloud points and the 79 categories contains all the 800 2D objects by putting the rare ones into the category of others. Detail list of those sub groups can be found in the appendix.\\

\textbf{simCrossTrans performance.} As shown in table \ref{2d_res}, with simCrossTrans we achieve $13.2\%$ and $16.1\%$ absolute performance gain (by evaluating AP50 based the SUNRGBD16 subgroup) compared with without simCrossTrana based on ConvNets and ViTs separately. This improvement is significant and beyond our expectations showing the effectiveness of using simCrossTrans. \\

\textbf{Compare with previous depth image only based 2D detection.} When comparing with the previous SOTA 3D image only system Frustum VoxNet \cite{Shen_2020_WACV}, our Swin-T based one achieve a significantly improvement by a large margin of $+15.4\%$ mAP50. This is promising and we give detailed analysis about the possible reason in the Discussion session.\\

\textbf{Compare with previous RGB or RGB-D image based 2D detection.} We further compare with our 3D point clouds based 2D detection system with SOTA RGB image based 2D detection system, there is only $1\%$ gap. It shows that power of using simCrossTrans. This is meaningful as we further reduce the gap between the 3D only image based detection system to the 2D RGB image based system. \\

\textbf{More results.} Besides the AP50, which was mainly used in previous works, we also use AP75 and AP to compare the results based on with simCrossTrans and without simCrossTrans. Meanwhile, we also report AP Across Scales of small, medium and large by following the same standard of COCO dataset. Those results can be found in Table \ref{more_2d_res}. From the results, we see that the withCrossTrans by either ConvNets based or ViT based, we can achieve significant performance improvements. At the same time ViT, specifically Swin-T based, can achieve a even better results with simCrossTrans. \\

\textbf{Model size and inference speed.} We compare the model size by number of parameters and the inference speed in Table \ref{infer_time_comp_}. The Swin-T based Mask-RCNN network has slightly more parameters compared to ResNet-50 based one. All the inference speed are tested based on a single Titan GPU. Although Swin-T and ResNet-50 has the similar number of FLOPs, the speed of ResNet-50 is much faster than Swin-T based. We believe this is due to the NVIDIA CUDA Deep Neural Network library (cuDNN)'s optimization on convolution operations. About 10 FPS should be enough for slow moving robotic to make real-time decision. If faster inference speed is needed, we can use more advanced GPU.\\
%\textbf{Results visualization:} In Figure \ref{fig:more_vis} we show some visualization of the 2D object detection. From the visualization, we can see that with simCrossTrans, the 2D detection performance \\

\subsection{Discussion}
\textbf{Why simCrossTrans works?} From the results shown in the Table \ref{2d_res} and \ref{more_2d_res} we can see the big influence for the simCrossTrans on both ConvNets and ViT based networks. In Figure \ref{fig:mAP_sunrgbd79}, we plot the mAP curve for the SUNRGBD79 categories to compare the performance of simCrossTrans. It shows clear difference of with and without simCrossTrans on the performance. In order to better understand why the simCrossTrans works so well, we plot the loss for different components related to the 2D object detection as shown in Figure \ref{fig:losses}. From the plots, we can see with the simCrossTrans, all component's loss are significantly reduced. We believe this is due to the pre-training based on the RGB images has well learned the context information need to better detect objects. For example, if we can detect the bed, then we will expect there is night stand or lamps near the bed.  Representations learned from contextual information was shown to be useful in NLP tasks as shown in BERT\cite{devlin-etal-2019-bert}. In \cite{Hao2019VisualizingAU} it claims: 1) pre-training leads to wider optima on the loss landscape, and eases optimization compared with training from scratch. 2) The pre-training-then-fine-tuning paradigm is robust to overfitting. From our observation shown in Figure \ref{fig:mAP_sunrgbd79}, by comparing of with simCrossTrans and without simCroassTrasn, we can see the time used to achieve the best performance is almost the same for both the ConvNets based and ViT based. We believe the pretraining can put the fine-tuning stage into an optimization landscape with much lower initial loss values in general. Meanwhile, in the fine-tune stage, the data used to train the model will add more details to the optimization landscape, which makes training from scratch and with pre-training take similar time to reach the (sub)optimal ending point. We did not observe the pre-training's improvement on overfitting. The deep understanding of why simCrossTrans works will be one of the future works.\\

\begin{table}[h]
\scriptsize
\begin{center}
               \resizebox{1.0\linewidth}{!}{
%   \resizebox{1.0\textwidth}{!}{
                               %\resizebox{\textwidth}{!}{
\begin{tabular}{|c|c|c|c|c|c|c|c|c|}
\hline

&\multicolumn{2}{c|}{image or block output shape}&\multicolumn{2}{c|}{receptive field}&\multicolumn{4}{c|}{channel/dimension}\\
  \multirow{2}{*}{}  & \multirow{2}{*}{ResNet-50}	&\multirow{2}{*}{Swin-T}&	\multirow{2}{*}{ResNet-50}	&\multirow{2}{*}{Swin-T}	&\multirow{2}{*}{ResNet-50}&	\multicolumn{3}{c|}{Swin-T}\\
  &&&&&&dim&head&head*dim\\	
\hline\hline
original image&	(530, 730, 3)&	(530, 730, 3)&&&&&&\\						
resized image&	(800, 1120, 3)&	(800, 1120, 3)&&&&&&\\							
Block 1 &	(200, 280, 256)&	(200, 280, 96)&	3 X 3&	28 X 28	&256&	96&	3&	288\\
Block 2	&(100, 140, 512)&	(100, 140, 192)&	3 X 3&	7 X 7&	512&	192	&6&	1152\\
Block 3	&(50, 70, 1024)&	(50, 70, 384)&	3 X 3&	7 X 7&	1024	&384&	12&	4608\\
Block 4	&(25, 35, 2048)	&(25, 35, 768)&	3 X 3&	7 X 7&	2048&	768&	24	&18432\\
\hline
\end{tabular}}
\end{center}
\caption {Details of image/tensor shape updates, receptive fields and the number of channel/dimension for ResNet-50 backbone and Swin-T backbone based on our experiment on SUN RGB-D dataset. For the Swin-T backbone, dim means the dimension of the related block's output. The head means the number of heads for the multi-head attention. }
\label{reception_fields}
\end{table}
\begin{figure}[t]
\begin{center}
   \includegraphics[width=1.0\linewidth]{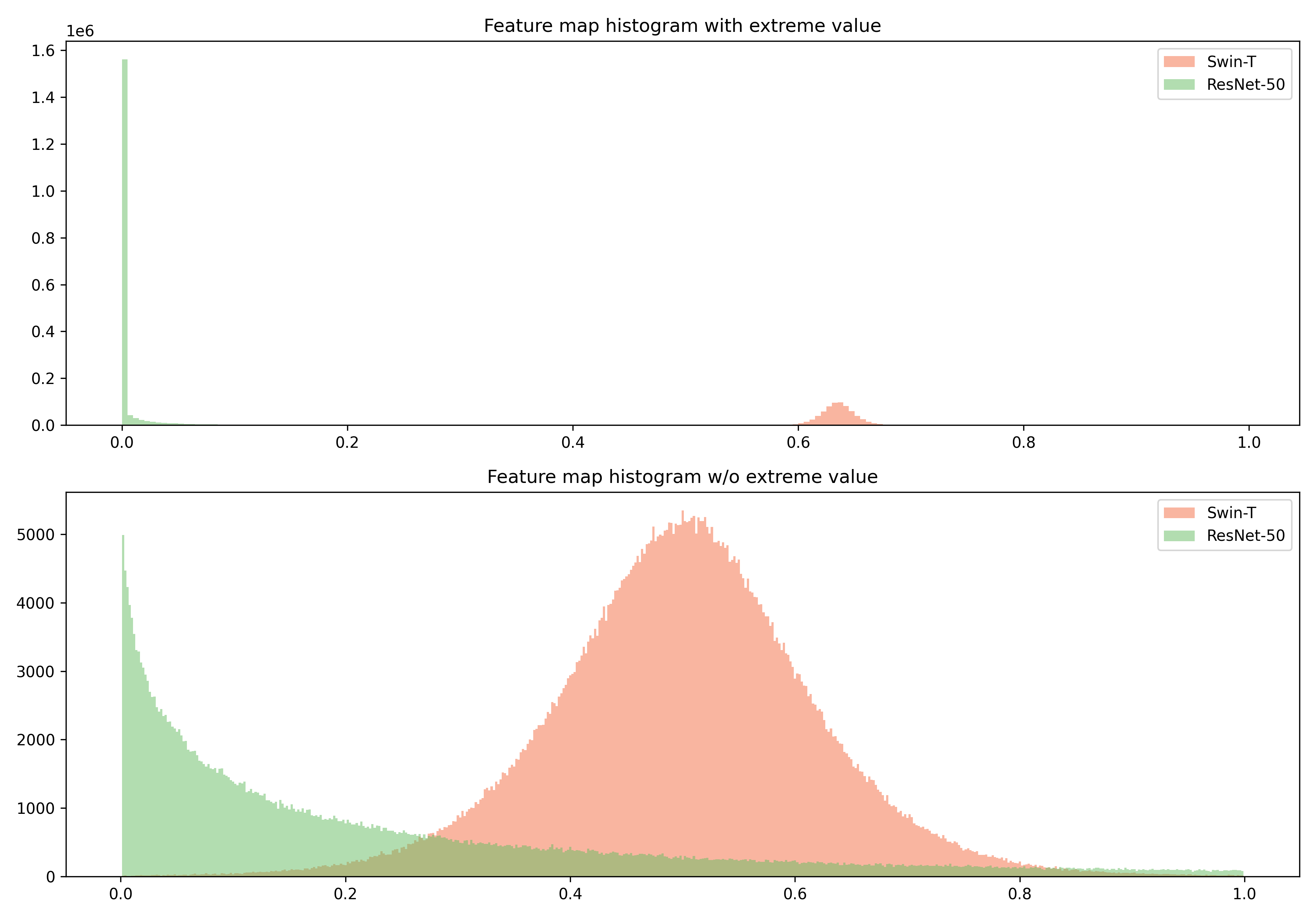}
\end{center}
    \caption{Histogram of the last layer feature map for ResNet-50 and Swin-T. The top one is the histogram of normalize all the values to 0 and 1 without removing the extreme values. The bottom one is the histogram of normalize all the values based on the top one by using the values in range [0, 0.1] for ResNet and [0.57, 0.7] for Swin-T.}
\label{fig:hist}
\end{figure}

\begin{figure*}[t]
\begin{center}
   \includegraphics[width=1.0\linewidth]{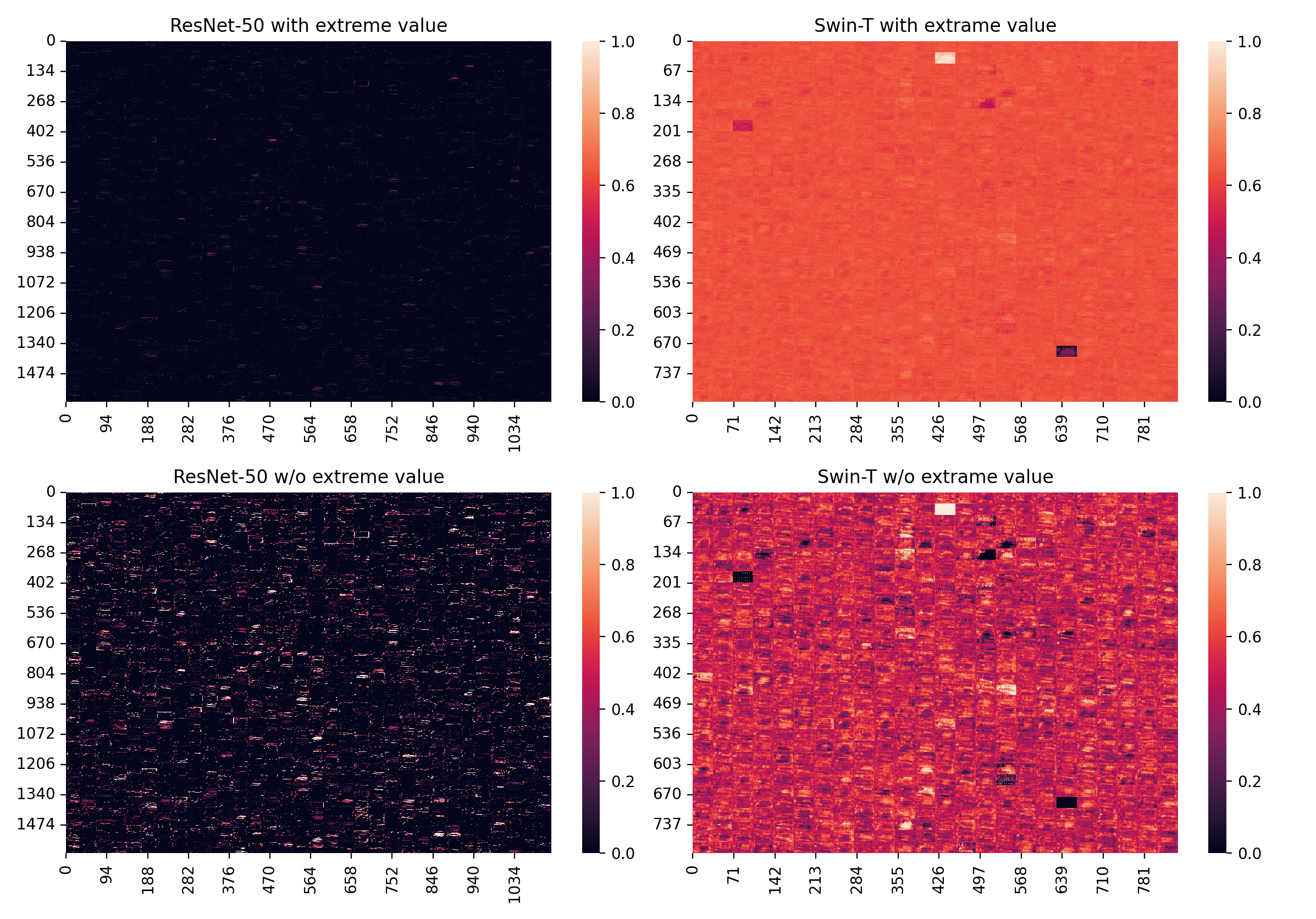}
\end{center}
    \caption{Combined feature maps of last layer backbone output. For ResNet, the last layer's original shape is (25, 35, 2048) and for Swin-T is	(25, 35, 768). We choose 64 by 32 grids for resnet and 32 by 24 grids for Swin-T and put each channel's feature map into one grid. }
\label{fig:feature}
\end{figure*}

\begin{figure}[t]
\begin{center}
   \includegraphics[width=1.0\linewidth]{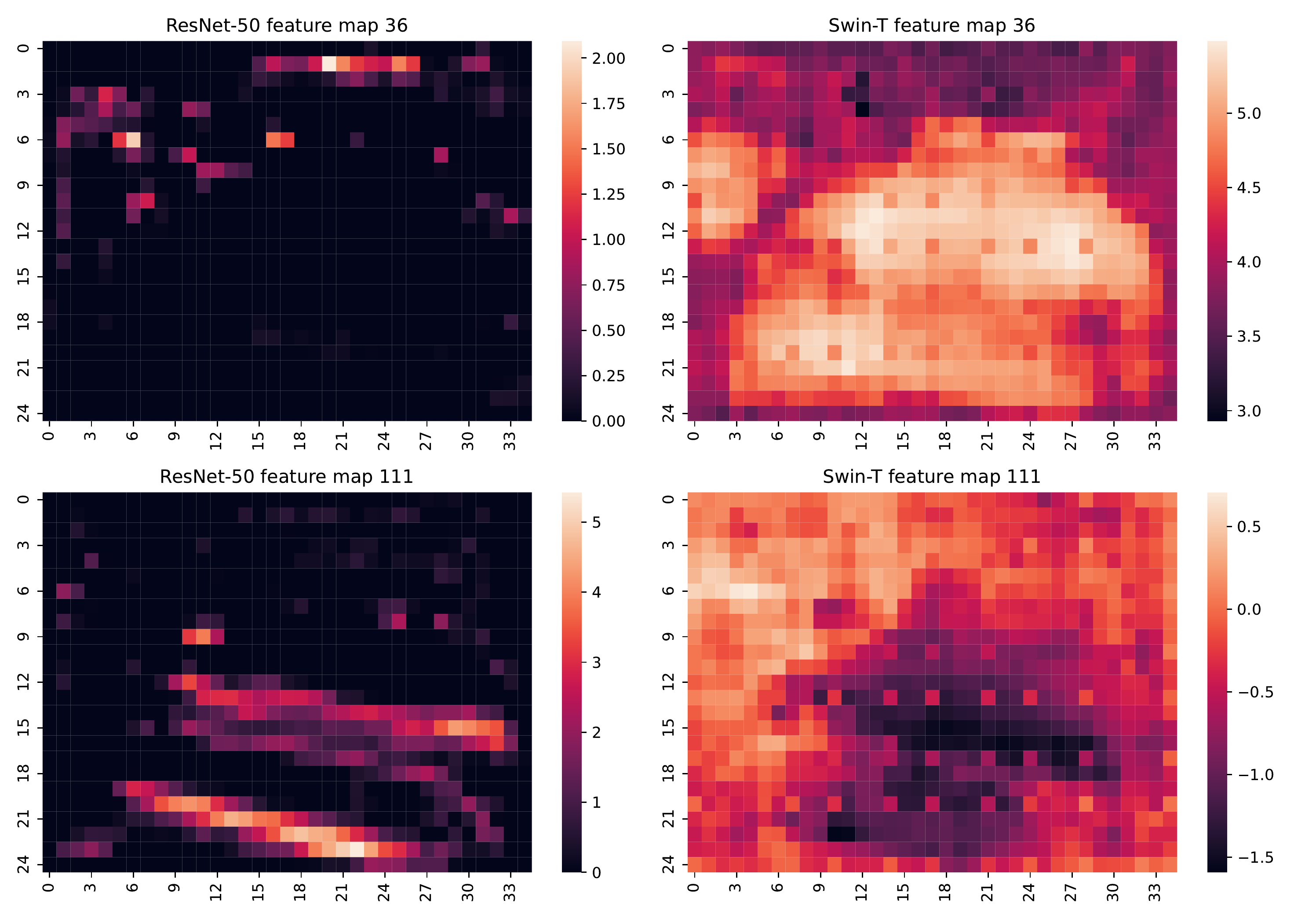}
\end{center}
    \caption{Feature maps of channel 36 and 111 from the backbone last layer output. Left two are from the ResNet and the rigth two are from the Swin-T.}
\label{fig:feature_each}
\end{figure}

\textbf{Why ViT is better than ConvNets?} From the results, we can see that the Swin Transformer\cite{liu2021Swin} based achieved a significant better result than ConvNet based on both with simCrossTrans and without simCrossTrans, although they have a similar computation complexity. In \cite{Cordonnier2020On}, the paper claims that a multi-head self-attention layer with sufficient number of heads is at least as expressive as any convolutional layer. In the ViT\cite{dosovitskiy2021an} paper, it claims that "Transformers lack some of the inductive biases inherent to CNNs, such as translation equivariance and locality, and therefore do not generalize well when trained on insufficient amounts of data. But for larger datasets, learning the relevant patterns directly from data is sufficient, even beneficial." From our observation, we believe the following are the critical reasons that Swin-T performs better than ResNet-50: \\
\begin{equation}
    FLOPs = 2*H*W (C_{in}*K^2 + 1)*C_{out}
\end{equation}
\begin{itemize}
\item A better global view: Based on transformer \cite{transformer_cite}, a self-attention layer connects all positions with a constant number of sequentially executed operations, whereas a ConvNet layer requires $O(\log_k^n)$ operations, where $k$ is the kernel size and $n$ is the sequence length (in vision, it is related to image width/height). Swin-T\cite{liu2021Swin} applies transformer in each small window, which has $M X M$ image patches. The $M$ is typical set as 7. It means that the receptive field for Swin-T is in a $7X7$ scale. Meanwhile, at the first input layer, the image are organized by each patch, each patch contain $4X4$ raw pixels images, so the effective field of the first layer is $28 x 28$, which is pretty large compared with ConvNet kernel which has a typical size of $3x3$. What's more, the shifted windowing introduced in Swin Transformer \cite{liu2021Swin} further increase the receptive field. The shifted windowing was proved effective in object detection in the Swin Transformer work. Increasing the kernel size of ConvNet can solve the problem, however, the increased kernel size will greatly increase the FLOPs as it is increasing quadratically with the kernel size (see equation 1 from \cite{DBLP:journals/corr/MolchanovTKAK16}). A detailed comparison of the receptive field of ResNet-50 and Swin-T based on the SUN RGB-D dataset we used is shown in Table \ref{reception_fields}. 
\item Swin-T has more powerful feature extraction unit: As shown in Table \ref{reception_fields}, when comparing the number of channel/dimensions used for ResNet-50 and Swin-T, we firstly see that ResNet-50 has more number of channels. However, when taking  the number of multi-head attention head number into consideration, the Swin-T has more basic computation units.
\end{itemize}

\begin{figure*}[t]
\begin{center}
   \includegraphics[width=1.0\linewidth]{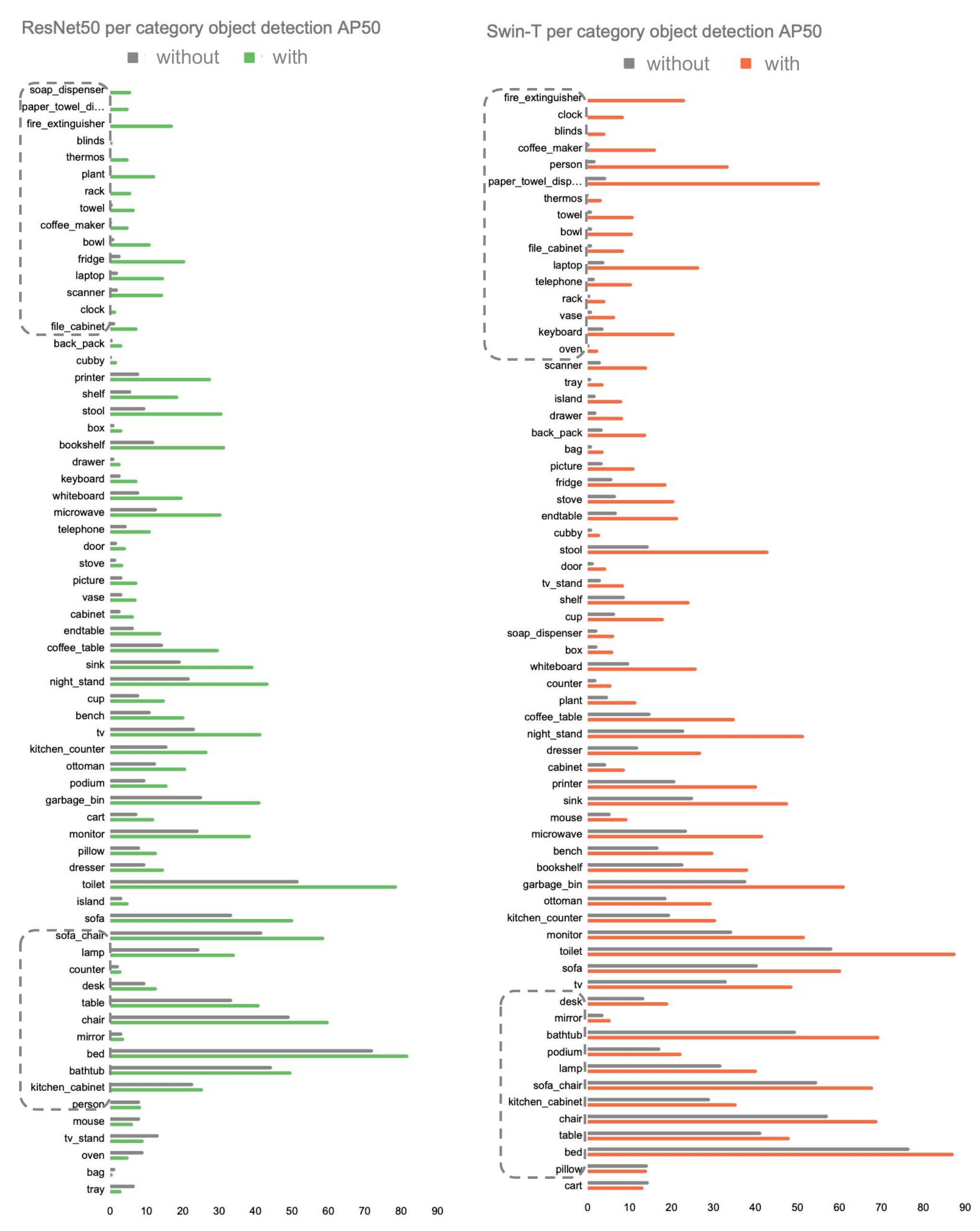}
\end{center}
    \caption{Performance of simCrossTrans on different categories. The categories are sorted by the performance improvement ratio of with simCrossTrans vs without. For each one, the top dashed block categories' improvement ratio: $[5, \infty]$; The bottom dashed block: $[1, 1.5]$. Between two blocks: $[1, 1.5]$; Below the bottom one: $[0, 1]$. }
\label{fig:with_without0}
\end{figure*}

Another two important differences between Swin-Transformer to ResNet are: 1) Replacing the pooling operations to Patch merging. 2) Using layer normalization (LN) \cite{LN_cite} instead of batch normalization (BN)\cite{BN_cite}. In \cite{convnet2020s}, the auther replacing the BN in ConvNets with LN and did not observe significant performance difference. We are not clear about the influence of those two to the performance. It can be further explored. We plot the histogram of last layer's feature map and also visualize the last layer's feature map output from both the ResNet-50 and Swin-T from our experiments based on using simCrossTrans ones. See Figure \ref{fig:hist} and \ref{fig:feature}. From Figure \ref{fig:hist}, we can see the clear difference of the last layer's feature map value distribution. For the Swin-T one, the distribution follow a bell shape due to the layer normalization \cite{LN_cite}.  In Figure \ref{fig:feature}, we visualize the last layer backbone's feature map. From the visualization we can clear see that ResNet has a more sparse outputs due to the pooling layers and not using layer normalization. The Swin-T's last layer is more dense. Also, it is interesting to see that for the Swin-T, there is a clear feature map which is similar to the original input image. We thought this might be due to the residual connection in both the ResNet and transformer, however, we did not find similar ones in the ResNet outputs. Figure \ref{feature_each} shows more details about 2 channel feature maps from ResNet and Swin-T.\\
\textbf{The performance of simCrossTrans on different categories}\\
In Figure \ref{fig:with_without0}, we show the performance improvement ratio of with simCrossTrans over without simCrossTrans. From the result, we can see that for easy categories such as bed, sofa chair and bathtub, with simCrossTrans helps. Since the signals of those easy categories are strong enough, the system without simCrossTrans also have a comparable performance. However, for the hard categories, such as laptop, keyboard and stool, with simCrossTrans, the performance is greatly improved. Which shows that the contextual knowledge learned from other modality data are helpful for detecting those hard objects with depth sensor only.\\

\section{Conclusion and future directions}
We introduce a simple but efficient cross sensor/modal transfer learning. Instead of using different encoder to extract the information from different modal, we unified the data representation between different modal to build an easy to implement approach, which is simCrossTrans. This simple approach works surprisingly well: by using simCrossTrans, without significantly modifying the model architecture, we can fine-tune the model based on learned weights from other modal and achieve much better performance than train the model from scratch, especially when combing with vision transformers. \\
%We hope that our one small step on inter modal transferring learning can contribute on the multi-modal learning, which is more close to human being's learning process.\\

We are excited about the good performance of our approach. This approach can greatly reduce the amount of data needed from expensive or not used by the general public sensors when training a model. Meanwhile, we are think about the future directions of extending our current approach: First, the easy implementation of our approach can help our model's performance improving together with the upgrading of the pre-training models. For example, 
%can we use a pre-trained model based on the self supervised approach such as MAE \cite{DBLP:journals/corr/abs-2111-06377} to see the performance? 
can we try the larger models, such as Swin Transformer v2 \cite{DBLP:journals/corr/abs-2111-09883} which has up to 3 billion parameters and push the COCO's detection performance from $50\%$, to  $63.1\%$ update the result (unfortunately, the Swin Transformer v2 needs Nvidia A100-40G GPUs, which are not available in our lab yet)? Second, in this article, we fine tune a new dataset from the pre-trained model only based depth image. One possible feature direction can be fine tuning the pre-trained model on both RGB and depth image based on a same model to make it work for both RGB and pseudo image generated from point cloud. By doing this, changing the model architecture and updating the model weights are not needed when changing the input data modal. This kind of system can work in such scenario: a robotic can run this unified model by taking the RGB data at the day time and using the depth sensor data at the nighttime without any adjustment to the model. Third, the transfer learning used here is based on supervised pre-training. Pre-train a model by using the Self supervised approach such as \cite{DBLP:journals/corr/abs-2111-06377} based on one modal and then fine tune the model in another modal will be another interesting direction to investigate. Limited to our knowledge, we can not list all the possibilities and we hope our proposed cross sensor transfer learning approach and our finding of combining simCrossTrans with vision transformers will inspire more future works.\\
\section{Acknowledgement}
We thank Zhujun Li to prepare the Titan GPU server used for this work. We also would like to thank Katharine Shen for proof reading, helpful comments and advice. We thank Izzy for showing up in our Figure \ref{fig:pipeline}.\\
{\small
\bibliographystyle{ieee_fullname}
\bibliography{egbib}
}

\end{document}